\documentclass[twoside,11pt]{article}

\usepackage{jmlr2e}
\usepackage{amsmath}
\usepackage{booktabs}
\usepackage{lastpage}

\hypersetup{
    colorlinks,
    breaklinks,
    urlcolor = black,
    linkcolor = blue,
    citecolor = blue,
}

\newcommand{\R}{\mathbb{R}}

\newcommand{\W}{\mathcal{W}}
\newcommand{\w}{\mathbf{w}}

\newcommand{\bbeta}{\boldsymbol{\beta}}
\newcommand{\Reg}{\operatorname{Regret}}
\newcommand{\SAReg}{\operatorname{SA-Regret}}
\newcommand{\DReg}{\operatorname{D-Regret}}

\newtheorem{assumption}{Assumption}

\ShortHeadings{Strongly Adaptive Online Learning Achieves Uniform Discounted Regret}{Yang, Yang and Zhang}
\firstpageno{1}

\begin{document}

\title{Strongly Adaptive Online Learning Achieves\\ Uniform Discounted Regret}

\author{\name Sifan Yang \email yangsf@lamda.nju.edu.cn \\
\name Wenhao Yang \email yangwh@lamda.nju.edu.cn \\
    \name Lijun Zhang \email zhanglj@lamda.nju.edu.cn \\
       \addr State Key Laboratory of Novel Software Technology, Nanjing University, Nanjing, China\\
School of Artificial Intelligence, Nanjing University, Nanjing, China
}

\maketitle

\begin{abstract}%
Discounted regret measures the performance of an online learner by assigning exponentially decaying weights to past losses, reflecting the greater importance of recent losses in non-stationary online learning. While previous work obtains uniform discounted regret bounds over a continuous range of discount factors using a specialized meta-algorithm, its results are limited to convex functions. In this paper, we demonstrate that strongly adaptive regret bounds can \emph{directly imply} uniform discounted regret guarantees for convex, strongly convex, and exp-concave loss functions. Our key observation is that discounted regret against a fixed comparator admits an exact decomposition into a convex combination of interval regrets. For convex functions, this reduction yields an $O(\sqrt{\log T/(1-\lambda)})$ discounted regret bound for all discount factors $\lambda\in(0,1)$, matching the previous rate over a broader range of discount factors. For strongly convex and exp-concave functions, we provide regret bounds of $O\left(\log T\left(1+\log\frac{1}{1-\lambda}\right)\right)$.
\end{abstract}

\begin{keywords}
online convex optimization, discounted regret, strongly adaptive regret, adaptive regret, exp-concave functions
\end{keywords}

\section{Introduction}\label{sec:intro}

Online convex optimization (OCO) is a classic framework for sequential decision making \citep{zinkevich2003online,hazan2022introduction}, where a learner chooses a decision $\w_t$ from a convex domain $\W\subseteq\R^d$ in each round $t\in[T]$ and then suffers a convex loss $f_t(\w_t)$ selected by the environment. The performance of the learner is commonly measured by the \emph{static regret} \citep{zinkevich2003online}, which compares its cumulative loss with that of the best fixed decision in hindsight.  In non-stationary environments, recent losses are often more relevant than earlier ones, whereas static regret assigns equal weight to all rounds. To account for this difference, the $\lambda$-discounted regret \citep{cesabianchi2006prediction,freund2008new} assigns a weight of $\lambda^{T-t}$ to the loss incurred at round $t$ and is defined as
\begin{equation*}
\DReg(T,\lambda)=\sum_{t=1}^{T}\lambda^{T-t}f_t(\w_t)-\min_{\w\in\W}\sum_{t=1}^{T}\lambda^{T-t}f_t(\w),
\end{equation*}
where $\lambda\in(0,1)$ controls the rate of discounting. When $\lambda=1$, discounted regret reduces to static regret. 
Discounted regret has been studied in adaptive regularization \citep{zhang2024discounted}, the analysis of Adam \citep{ahn2024understanding}, and online linear regression \citep{jacobsen2024online}. For a given discount factor, online gradient descent (OGD) with a step size proportional to $\sqrt{1-\lambda}$ achieves an $O(1/\sqrt{1-\lambda})$ discounted regret bound. \citet{yang2026discounted} further establish an $O(\sqrt{\log T/(1-\lambda)})$ discounted regret bound simultaneously for all discount factors in a continuous interval using a single algorithm.

The appropriate discount factor depends on how fast the environment changes, which is rarely known in advance. It is desirable to obtain discounted regret bounds simultaneously for all discount factors using a single algorithm. Directly aggregating OGD instances tuned to different discount factors is challenging, since each instance is analyzed under its own discounted regret, whereas standard meta-algorithms such as Hedge compare experts under a common performance measure. SOGD \citep{yang2026discounted} addresses this challenge by aggregating OGD instances tuned to a geometric grid of discount factors through the Discounted-Normal-Predictor with conservative updating (DNP-cu) \citep{kapralov2010prediction}, a carefully designed meta-algorithm. Discounted regret assigns larger weights to more recent rounds, and regret over suffix intervals captures the learner's performance on these rounds. For any interval $[r,s]\subseteq[T]$, the interval regret is defined as
\begin{equation*}
\Reg([r,s])=\sum_{t=r}^{s}f_t(\w_t)-\min_{\w\in\W}\sum_{t=r}^{s}f_t(\w).
\end{equation*}
Adaptive regret \citep{hazan2007adaptive} measures the maximum regret
over all contiguous intervals, while strongly adaptive regret
\citep{daniely2015strongly} additionally accounts for the interval
length $\tau$ and is defined as
\begin{equation*}
\SAReg(T,\tau)=\max_{1\le r\le T-\tau+1}\Reg([r,r+\tau-1]).
\end{equation*}
Discounted regret assigns larger weights to more recent rounds, while strongly adaptive regret controls regret over any contiguous interval, including intervals ending at the final round $T$. This paper investigates whether  \emph{interval regret guarantees can establish uniform discounted regret bounds.}

\begin{table}[t]
\centering
\caption{Uniform discounted regret bounds for strongly adaptive algorithms. All bounds hold simultaneously for $\lambda\in(0,1)$.}
\label{tab:summary}
\small
\renewcommand{\arraystretch}{1.3}
\setlength{\tabcolsep}{4pt}
\resizebox{\linewidth}{!}{
\begin{tabular}{@{}cccc@{}}
\toprule
Functions & Algorithm & $\DReg(T,\lambda)$ & Active experts \\
\midrule
Convex
& \shortstack[c]{CBCE\\ \citep{jun2017improved}}
& $O\left(\sqrt{\frac{\log T}{1-\lambda}}\right)$
& $O(\log(t+1))$ \\
\addlinespace[3pt]

Strongly convex
& \shortstack[c]{IFLH ($K=2$)\\ \citep{zhang2018dynamic}}
& $O\left(\frac{\log T}{\mu}\left(1+\log\frac{1}{1-\lambda}\right)\right)$
& $O(\log(t+1))$ \\

Strongly convex
& \shortstack[c]{IFLH ($K=\lceil T^{1/q}\rceil$)\\ \citep{zhang2018dynamic}}
& $O\left(\frac{q\log T}{\mu}\right)$
& $O(qT^{1/q})$ \\
\addlinespace[3pt]

Exp-concave
& \shortstack[c]{AFLH\\ \citep{hazan2007adaptive}}
& $O\left(\frac{d\log T}{\alpha}\left(1+\log\frac{1}{1-\lambda}\right)\right)$
& $O(\log(t+1))$ \\

Exp-concave
& \shortstack[c]{FLH\\ \citep{hazan2007adaptive}}
& $O\left(\frac{1}{\alpha}\left(d\left(1+\log\frac{1}{1-\lambda}\right)+\log T\right)\right)$
& $t$ \\
\bottomrule
\end{tabular}
}

\vspace{1mm}
\begin{minipage}{\linewidth}
\footnotesize
Here $q\ge2$ is a fixed integer, and $T\ge2$.
For exp-concave functions, we use a valid
exp-concavity parameter satisfying $\alpha GD\le1$.
The last column reports the number of active experts at round $t$.
\end{minipage}
\end{table}

For any nonincreasing discount sequence, the discounted regret against a fixed comparator can be expressed as a convex combination of regrets over intervals of $[r,T]$. Since strongly adaptive regret controls the regret over every interval, this decomposition provides a direct reduction from strongly adaptive regret to discounted regret. In particular, if an algorithm admits a strongly adaptive regret bound $\psi_T(\tau)$ that is concave in the interval length $\tau$, it achieves
\begin{equation*}
\DReg(T,\lambda)\leq\psi_T(\tau_\lambda),
\end{equation*}
simultaneously for all $\lambda\in(0,1)$, where $\tau_\lambda=\sum_{k=0}^{T-1}\lambda^k\leq 1/(1-\lambda)$.

Existing strongly adaptive algorithms achieve uniform discounted regret bounds through this reduction, as summarized in Table~\ref{tab:summary}. For convex functions, CBCE \citep{jun2017improved} achieves an $O(\sqrt{\log T/(1-\lambda)})$ discounted regret bound for all $\lambda\in(0,1)$, matching the rate of SOGD \citep{yang2026discounted} over a broader range of discount factors. For strongly convex and exp-concave functions, IFLH \citep{zhang2018dynamic} and AFLH \citep{hazan2007adaptive} achieve $O\left(\log T\left(1+\log\frac{1}{1-\lambda}\right)\right)$ discounted regret bounds. These algorithms maintain only $O(\log t)$ experts at round $t$, while FLH \citep{hazan2007adaptive} achieves sharper regret bounds at the cost of $t$ active experts.

\paragraph{Organization.} In the rest of the paper, Section~\ref{sec:related} reviews the related work, Section~\ref{sec:setup} introduces the necessary assumptions, Section~\ref{sec:reduction} establishes the reduction from strongly adaptive regret to discounted regret, Section~\ref{sec:consequences} derives uniform discounted regret bounds for existing strongly adaptive methods, and Section~\ref{sec:conclusion} concludes this paper.

\section{Related Work}\label{sec:related}

\subsection{Adaptive and Strongly Adaptive Regret}

Static regret evaluates the performance of an online learner against a fixed decision over all rounds, which may fail to capture their performance in changing environments. To address this issue, \citet{hazan2007adaptive} introduce adaptive regret, which measures the maximum static regret over all contiguous intervals $[r,s]\subseteq[T]$. They establish an $O(\log T)$ adaptive regret bound for exp-concave functions, but their algorithm requires $T$ active experts. They further develop a more efficient algorithm at the cost of a worse regret bound.

\citet{daniely2015strongly} introduce strongly adaptive regret, which measures the maximum regret over intervals of a given length $\tau$, rather than over all intervals as in adaptive regret. For convex functions, their algorithm achieves an $O(\sqrt{\tau}\log T)$ regret bound. \citet{jun2017improved} improve this bound to $O(\sqrt{\tau\log T})$ by using coin betting. For strongly convex and exp-concave functions, \citet{zhang2018dynamic} establish logarithmic strongly adaptive regret bounds with explicit dependence on the interval length $\tau$.

\subsection{Discounted Online Learning}
Discounted regret has been studied in prediction with expert advice. \citet{cesabianchi2006prediction} consider general nonincreasing discount sequences, while \citet{freund2008new} propose NormalHedge, which achieves an $O(\sqrt{\log N/(1-\lambda)})$ discounted regret bound against $N$ experts. In the field of OCO, \citet{zhang2024discounted} develop discounted FTRL with instance-adaptive regularization for time-varying discount factors, which are assumed to be known to the learner in advance.

\citet{zhang2022smoothed} propose smoothed OGD (SOGD), which aggregates OGD experts using the Discounted-Normal-Predictor (DNP) \citep{kapralov2010prediction}. Building on SOGD, \citet{yang2026discounted} obtain an $O(\sqrt{\log T/(1-\lambda)})$ discounted regret bound simultaneously over a continuous interval of discount factors without knowing $\lambda$ in advance. However, their approach relies on a specialized meta-algorithm, and the regret guarantee is limited to convex functions.

\section{Preliminaries}\label{sec:setup}
\paragraph{Notation.} For integers $1\le r\le s\le T$, we denote
$[r,s]=\{r,r+1,\ldots,s\}$ and $[T]=[1,T]$.
We use $\|\cdot\|$ and $\langle\cdot,\cdot\rangle$
to denote the Euclidean norm and inner product,
respectively.

We introduce the following standard assumptions \citep{zhang2022smoothed,yang2026discounted}.

\begin{assumption}\label{asm:convex}
Each loss function $f_t(\cdot)$ is convex on the feasible domain $\W$.
\end{assumption}

\begin{assumption}\label{asm:strong}
Each loss function $f_t(\cdot)$ is $\mu$-strongly convex on the feasible domain $\W$
for some $\mu>0$, i.e., for any $\w,\w'\in\W$,
\begin{align*}
f_t(\w')
\ge f_t(\w)
+\langle\nabla f_t(\w),\w'-\w\rangle
+\frac{\mu}{2}\|\w'-\w\|^2.
\end{align*}
\end{assumption}

\begin{assumption}\label{asm:expconcave}
Each loss function $f_t(\cdot)$ is $\alpha$-exp-concave on the feasible domain $\W$
for some $\alpha>0$, i.e., $\exp(-\alpha f_t)$ is concave on the feasible domain $\W$.
\end{assumption}

\begin{assumption}\label{asm:gradient}
The gradients are bounded by $G>0$, i.e.,
$\|\nabla f_t(\w)\|\le G$ for all
$t\in[T]$ and $\w\in\W$.
\end{assumption}

\begin{assumption}\label{asm:domain}
The feasible domain $\W\subseteq\R^d$ is convex
with diameter at most $D>0$, i.e.,
$\|\w-\w'\|\le D$ for all $\w,\w'\in\W$.
\end{assumption}

Without loss of generality, we further assume that the losses satisfy
\begin{align*}
0\le f_t(\w)\le GD,
\quad \forall\,t\in[T],\ \w\in\W.
\end{align*}

Then we introduce the definitions of regrets. For any interval $[r,s]\subseteq[T]$ and comparator $\w\in\W$, we define the interval regret as
\begin{equation*}
    \Reg_{[r,s]}(\w)=\sum_{t=r}^{s}\big(f_t(\w_t)-f_t(\w)\big),
\qquad
\Reg([r,s])=\max_{\w\in\W}\Reg_{[r,s]}(\w),
\end{equation*}
where the second quantity compares the learner with the best fixed decision on $[r,s]$. Adaptive regret is the maximum of $\Reg([r,s])$ over all intervals \citep{hazan2007adaptive}, which is defined as 
\begin{equation*}
    \operatorname{A-Regret}(T)
=\max_{1\le r\le s\le T}\Reg([r,s]).
\end{equation*}
For a given interval length $\tau$, strongly adaptive regret \citep{daniely2015strongly} is defined as
\begin{equation*}
    \SAReg(T,\tau)=\max_{1\le r\le T-\tau+1}\Reg([r,r+\tau-1]),
\qquad \tau\in[T].
\end{equation*}
For discounted regret, we consider a nonincreasing discount sequence $\bbeta=(\beta_0,\ldots,\beta_{T-1})$ satisfying
\begin{equation}\label{eq:1}
1=\beta_0\ge\beta_1\ge\cdots\ge\beta_{T-1}\ge\beta_T=0.
\end{equation} 
The discounted regret against a fixed comparator $\w\in\W$ and the discounted regret of the learner are defined as
\begin{equation*}
\DReg_T^{\bbeta}(\w)
=\sum_{t=1}^{T}\beta_{T-t}\big(f_t(\w_t)-f_t(\w)\big),
\qquad
\DReg(T,\bbeta)=\max_{\w\in\W}\DReg_T^{\bbeta}(\w).   
\end{equation*}
This definition covers the general discounted regret \citep{cesabianchi2006prediction}. The special case $\beta_k=\lambda^k$ implies the $\lambda$-discounted regret \citep{freund2008new}, that is,
\begin{equation*}
    \DReg(T,\lambda)
=\sum_{t=1}^{T}\lambda^{T-t}f_t(\w_t)
-\min_{\w\in\W}\sum_{t=1}^{T}\lambda^{T-t}f_t(\w),
\end{equation*}
where $\lambda\in(0,1)$. In particular, $\lambda=0$ corresponds to the regret at the last round, while $\lambda=1$ recovers static regret.

We define the effective window size as the sum of the discount weights, i.e.,
\[
\tau_{\bbeta}=\sum_{k=0}^{T-1}\beta_k,\qquad
\tau_\lambda=\sum_{k=0}^{T-1}\lambda^k=
\begin{cases}
\dfrac{1-\lambda^T}{1-\lambda}, & \lambda\in(0,1),\\[2mm]
T, & \lambda=1.
\end{cases}
\]
By using \eqref{eq:1}, we have $1\le\tau_{\bbeta}\le T$. For exponential discounting, we further have
\begin{equation*}
1\le\tau_\lambda\le\min\left\{T,\frac{1}{1-\lambda}\right\},
\qquad \lambda\in(0,1).
\end{equation*}

For subsequent analysis, suppose an online algorithm admits a strongly adaptive regret bound $\psi_T(\cdot):[T]\to\R$ satisfying
\begin{equation}\label{eq}
\SAReg(T,\tau)\le\psi_T(\tau),
\qquad \forall \tau\in[T].
\end{equation}
The algorithm may depend on $T$, but does not require the discount sequence $\bbeta$ as input.

\section{From Strongly Adaptive Regret to Discounted Regret}\label{sec:reduction}

Strongly adaptive regret controls the cumulative regret over every contiguous interval, whereas discounted regret assigns different weights to individual rounds. To establish a connection between these two measures, the key challenge is to relate the weighted regret to the interval regrets controlled by strongly adaptive regret. An important property of discounted regret is that more recent rounds receive larger weights, which motivates considering intervals of the form $[r,T]$, $r\in[T]$. For any nonincreasing discount sequence, the discounted regret against a fixed comparator can be expressed as a convex combination of regrets over these intervals. Since strongly adaptive regret controls the regret on every interval, this decomposition provides a direct reduction from strongly adaptive regret to discounted regret.

\begin{lemma}\label{lem:suffix}
For any discount sequence $\bbeta$ satisfying
\eqref{eq:1} with $\beta_T=0$,
the discounted regret against any fixed comparator
$\w\in\W$ satisfies
\begin{equation}\label{eq:suffix}
\DReg_T^{\bbeta}(\w)
=\sum_{\tau=1}^{T}
(\beta_{\tau-1}-\beta_\tau)
\Reg_{[T-\tau+1,T]}(\w).
\end{equation}
Moreover,
\begin{equation}\label{eq:weight-properties}
\begin{aligned}
&\beta_{\tau-1}-\beta_\tau\ge0, \forall\tau\in[T],\qquad  \sum_{\tau=1}^{T}
\tau(\beta_{\tau-1}-\beta_\tau)
=\tau_{\bbeta}.
\end{aligned}
\end{equation}
\end{lemma}

\begin{proof}
Since $\beta_T=0$, for every $t\in[T]$, we have
\begin{align*}
\beta_{T-t}
&=\beta_{T-t}-\beta_T=\sum_{\tau=T-t+1}^{T}
(\beta_{\tau-1}-\beta_\tau).
\end{align*}
Substituting this identity into
$\DReg_T^{\bbeta}(\w)$, we obtain
\begin{align*}
\DReg_T^{\bbeta}(\w)
&=\sum_{t=1}^{T}\beta_{T-t}
\big(f_t(\w_t)-f_t(\w)\big)\\
&=\sum_{t=1}^{T}
\sum_{\tau=T-t+1}^{T}
(\beta_{\tau-1}-\beta_\tau)
\big(f_t(\w_t)-f_t(\w)\big)\\
&=\sum_{\tau=1}^{T}
(\beta_{\tau-1}-\beta_\tau)
\sum_{t=T-\tau+1}^{T}
\big(f_t(\w_t)-f_t(\w)\big)\\
&=\sum_{\tau=1}^{T}
(\beta_{\tau-1}-\beta_\tau)
\Reg_{[T-\tau+1,T]}(\w).
\end{align*}

Furthermore, we have
\begin{align*}
\sum_{\tau=1}^{T}
(\beta_{\tau-1}-\beta_\tau)
&=\beta_0-\beta_T =1.
\end{align*}
Finally, we obtain
\begin{align*}
\sum_{\tau=1}^{T}
\tau(\beta_{\tau-1}-\beta_\tau)&=\sum_{\tau=1}^{T}\tau\beta_{\tau-1}
-\sum_{\tau=1}^{T}\tau\beta_\tau\\
&=\beta_0
+\sum_{\tau=1}^{T-1}(\tau+1)\beta_\tau
-\sum_{\tau=1}^{T-1}\tau\beta_\tau
-T\beta_T\\
&=\sum_{\tau=0}^{T-1}\beta_\tau=\tau_{\bbeta}.
\end{align*}
\end{proof}
Lemma~\ref{lem:suffix} establishes a connection between discounted regret and interval regret. Since each interval regret is bounded by strongly adaptive regret at the corresponding length, the decomposition allows us to bound discounted regret using strongly adaptive regret guarantees. Moreover, the weighted average of the interval lengths equals $\tau_{\bbeta}$, which enables a tighter bound when the strongly adaptive regret bound is concave. The following theorem formalizes this reduction.

\begin{theorem}\label{thm:main}
Suppose an online algorithm satisfies
$\SAReg(T,\tau)\le\psi_T(\tau)$ for all $\tau\in[T]$.
For any discount sequence $\bbeta$ satisfying \eqref{eq:1}, we have
\begin{equation}\label{eq:main-exact}
\DReg(T,\bbeta)\le\sum_{\tau=1}^{T}(\beta_{\tau-1}-\beta_\tau)\psi_T(\tau).
\end{equation}
If $\psi_T$ admits a concave extension to $[1,T]$, then
\begin{equation}\label{eq:main-concave}
\DReg(T,\bbeta)\le\psi_T(\tau_{\bbeta}).
\end{equation}
Both bounds hold simultaneously for all discount sequences satisfying \eqref{eq:1}.
\end{theorem}

\begin{proof}
For any $\w\in\W$ and $\tau\in[T]$, we have
\begin{align*}
\Reg_{[T-\tau+1,T]}(\w)
&\le\Reg([T-\tau+1,T])\le\SAReg(T,\tau)\le\psi_T(\tau).
\end{align*}
Combining Lemma~\ref{lem:suffix} with the nonnegativity of
$\beta_{\tau-1}-\beta_\tau$, we can derive
\begin{align*}
\DReg(T,\bbeta)
&=\max_{\w\in\W}\sum_{\tau=1}^{T}(\beta_{\tau-1}-\beta_\tau)
\Reg_{[T-\tau+1,T]}(\w)\\
&\le\sum_{\tau=1}^{T}(\beta_{\tau-1}-\beta_\tau)
\max_{\w\in\W}\Reg_{[T-\tau+1,T]}(\w)\\
&=\sum_{\tau=1}^{T}(\beta_{\tau-1}-\beta_\tau)
\Reg([T-\tau+1,T])\\
&\le\sum_{\tau=1}^{T}(\beta_{\tau-1}-\beta_\tau)
\SAReg(T,\tau)\\
&\le\sum_{\tau=1}^{T}(\beta_{\tau-1}-\beta_\tau)\psi_T(\tau).
\end{align*}

If $\psi_T$ is concave on $[1,T]$, Jensen's inequality implies
\begin{align*}
\DReg(T,\bbeta)
&\le\sum_{\tau=1}^{T}(\beta_{\tau-1}-\beta_\tau)\psi_T(\tau)\\
&\le\psi_T\left(\sum_{\tau=1}^{T}
\tau(\beta_{\tau-1}-\beta_\tau)\right)\\
&=\psi_T(\tau_{\bbeta}),
\end{align*}
where the second inequality uses
$\sum_{\tau=1}^{T}(\beta_{\tau-1}-\beta_\tau)=1$.
\end{proof}

\begin{remark}\label{rem:reduction}
The decomposition in Lemma~\ref{lem:suffix} holds for a fixed comparator, whereas the maximization over comparators in discounted regret leads to the inequality in \eqref{eq:main-exact}. More importantly, Theorem~\ref{thm:main} transfers the strongly adaptive regret guarantee to all nonincreasing discount sequences simultaneously. This reduction does not require convexity or other assumptions on the loss functions, which are only needed to establish specific strongly adaptive regret bounds.
\end{remark}

For exponential discounting, the discount sequence takes the form $\beta_k=\lambda^k$ and we can derive the following corollary.

\begin{corollary}\label{cor:uniform}
Suppose an online algorithm satisfies $\SAReg(T,\tau)\le\psi_T(\tau)$ for all $\tau\in[T]$. Then, for all $\lambda\in(0,1)$ simultaneously,
\begin{equation*}
\DReg(T,\lambda)\le(1-\lambda)\sum_{\tau=1}^{T-1}\lambda^{\tau-1}\psi_T(\tau)+\lambda^{T-1}\psi_T(T).
\end{equation*}
If $\psi_T$ admits a concave extension to $[1,T]$, then
\begin{equation*}
\DReg(T,\lambda)\le\psi_T(\tau_\lambda).
\end{equation*}
\end{corollary}

\begin{proof}
For any $\lambda\in(0,1)$, setting $\beta_k=\lambda^k$ for $k=0,\ldots,T-1$ and $\beta_T=0$, we have
\begin{align*}
\beta_{\tau-1}-\beta_\tau&=(1-\lambda)\lambda^{\tau-1},\quad \tau\in[T-1],\\
\beta_{T-1}-\beta_T&=\lambda^{T-1}.
\end{align*}
Substituting these expressions into \eqref{eq:main-exact}, we obtain
\begin{align*}
\DReg(T,\lambda)
&=\DReg(T,\bbeta)\le\sum_{\tau=1}^{T}(\beta_{\tau-1}-\beta_\tau)\psi_T(\tau)\\
&=(1-\lambda)\sum_{\tau=1}^{T-1}\lambda^{\tau-1}\psi_T(\tau)+\lambda^{T-1}\psi_T(T).
\end{align*}
If $\psi_T$ admits a concave extension to $[1,T]$, we further have
\begin{align*}
\DReg(T,\lambda)
&\le\psi_T(\tau_{\bbeta})
=\psi_T\left(\sum_{k=0}^{T-1}\lambda^k\right)
=\psi_T(\tau_\lambda).
\end{align*}
\end{proof}

Theorem~\ref{thm:main} and Corollary~\ref{cor:uniform} establish a general reduction from strongly adaptive regret to discounted regret for nonincreasing discount sequences, including exponential discounting with any $\lambda\in(0,1)$. Since this reduction does not depend on the curvature of the loss functions, it can be applied to strongly adaptive regret bounds for different function classes. In the next section, we derive uniform discounted regret bounds for convex, strongly convex, and exp-concave functions.

\section{Discounted Regret of Strongly Adaptive Methods}\label{sec:consequences}

In this section, we apply our reduction to existing strongly adaptive algorithms and derive uniform discounted regret bounds for convex, strongly convex, and exp-concave functions, respectively.
\subsection{Convex Functions}

For convex functions, \citet{jun2017improved} propose CBCE, which uses OGD as the base learner and maintains $O(\log t)$ active experts at round $t$ through geometric covering intervals. Under Assumptions~\ref{asm:convex}, \ref{asm:gradient} and \ref{asm:domain}, their method achieves the following strongly adaptive regret bound for any interval $[r,s]\subseteq[T]$
\begin{equation*}
    \Reg([r,s])\le GD\left(\frac{6}{\sqrt{2}-1}+8\sqrt{7\log T+5}\right)\sqrt{s-r+1}.
\end{equation*}
Since the regret bound is concave in the interval length, Corollary~\ref{cor:uniform} implies that, simultaneously for all $\lambda\in(0,1)$,
\begin{align*}
\DReg(T,\lambda)
&\le GD\left(\frac{6}{\sqrt{2}-1}+8\sqrt{7\log T+5}\right)\sqrt{\tau_\lambda}=O\left(\sqrt{\tau_\lambda\log T}\right)=O\left(GD\sqrt{\frac{\log T}{1-\lambda}}\right),
\end{align*}
which matches the regret bound of \citet{yang2026discounted} over the range of discount factors they consider and extends the guarantee to all $\lambda\in(0,1)$.

\subsection{Strongly Convex Functions}\label{sec:iflh}

For strongly convex functions, \citet{zhang2018dynamic} propose improved following the leading history (IFLH), which uses OGD as the base learner and adopts a base-$K$ lifetime schedule to control the number of active experts. For any integer $K\ge 2$, IFLH maintains at most $(K-1)(\lfloor\log_K t\rfloor+1)$ active experts at round $t$. Under Assumptions~\ref{asm:strong}, \ref{asm:gradient}, and~\ref{asm:domain}, their algorithm achieves the following strongly adaptive regret bound:
\begin{equation*}
    \SAReg(T,\tau)\le\frac{G^2}{2\mu}\left[(1+3\log T)(\lceil\log_K\tau\rceil+1)+4\log T\right],\quad \forall\tau\in[T].
\end{equation*}

Since $\lceil\log_K\tau\rceil+1\le 2+\log\tau/\log K$, the above bound admits a concave upper bound in $\tau$. Combining Corollary~\ref{cor:uniform} with $\sum_{k=0}^{T-1}\lambda^k\le 1/(1-\lambda)$, we obtain
\begin{equation*}
    \DReg(T,\lambda)\le\frac{G^2}{2\mu}\left[2+10\log T+\frac{1+3\log T}{\log K}\log\frac{1}{1-\lambda}\right],
\end{equation*}
simultaneously for all $\lambda\in(0,1)$.

The parameter $K$ controls the trade-off between the regret bound and the number of active experts. When $K=2$, IFLH maintains $O(\log t)$ active experts and achieves
\begin{equation*}
    \DReg(T,\lambda)=O\left(\frac{G^2}{\mu}\log T\left(1+\log\frac{1}{1-\lambda}\right)\right).
\end{equation*}
Alternatively, for any fixed integer $q\ge2$, choosing $K=\lceil T^{1/q}\rceil$ ensures $\lceil\log_K\tau\rceil\le q$ for all $\tau\in[T]$. The strongly adaptive regret bound of IFLH and Corollary~\ref{cor:uniform} imply
\begin{align*}
    \DReg(T,\lambda)&\le\frac{G^2}{2\mu}\left[q+1+(3q+7)\log T\right]=O\left(\frac{qG^2}{\mu}\log T\right),
\end{align*}
with $O(qT^{1/q})$ active experts.

\subsection{Exp-Concave Functions}\label{sec:expconcave}

For exp-concave functions, \citet{hazan2007adaptive} develop FLH, which uses online Newton step (ONS) as the base learner and maintains $t$ active experts at round $t$. Under Assumptions~\ref{asm:expconcave}, \ref{asm:gradient}, and~\ref{asm:domain}, let $\gamma=\frac12\min\{\alpha,1/(4GD)\}$. Combining the ONS regret bound with the FLH aggregation guarantee, we have
\begin{equation*}
\SAReg(T,\tau)\le\frac{\log T}{\alpha}
+\frac1{2\gamma}\left[
1+d\log\left(1+\frac{\gamma^2G^2D^2}{d}\tau\right)
\right],\quad \forall\tau\in[T].
\end{equation*}
Since the right-hand side is concave in $\tau$, Corollary~\ref{cor:uniform} implies
\begin{align*}
\DReg(T,\lambda)
&\le\frac{\log T}{\alpha}
+\frac1{2\gamma}\left[
1+d\log\left(1+\frac{\gamma^2G^2D^2}{d(1-\lambda)}\right)
\right]\\
&=O\left(
d\left(\frac1\alpha+GD\right)
\left(1+\log\frac1{1-\lambda}\right)
+\frac{\log T}{\alpha}
\right),
\end{align*}
simultaneously for all $\lambda\in(0,1)$.

To reduce the computational cost of FLH, \citet{hazan2007adaptive} further develop AFLH, which maintains $O(\log t)$ active experts at the cost of a worse regret bound. With ONS as the base learner, AFLH achieves
\begin{equation*}
\SAReg(T,\tau)=O\left(
d\left(\frac1\alpha+GD\right)
\log T(1+\log\tau)
\right).
\end{equation*}
Applying Corollary~\ref{cor:uniform}, we obtain
\begin{equation*}
\DReg(T,\lambda)=O\left(
d\left(\frac1\alpha+GD\right)
\log T\left(1+\log\frac1{1-\lambda}\right)
\right),
\end{equation*}
simultaneously for all $\lambda\in(0,1)$.

To further control the trade-off between regret and computational cost, \citet{zhang2018dynamic} develop IFLH with a base-$K$ lifetime schedule. Using ONS as the base learner, IFLH maintains at most $(K-1)(\lfloor\log_K t\rfloor+1)$ active experts at round $t$ and achieves
\begin{equation*}
\SAReg(T,\tau)\le
\left[
\frac{(5d+1)(\lceil\log_K\tau\rceil+1)+2}{\alpha}
+5dGD(\lceil\log_K\tau\rceil+1)
\right]\log T,\quad \forall\tau\in[T].
\end{equation*}
Since $\lceil\log_K\tau\rceil+1\le2+\log\tau/\log K$, the above bound admits a concave upper bound in $\tau$. Corollary~\ref{cor:uniform} implies
\begin{equation*}
\DReg(T,\lambda)\le
\left[
\left(\frac{5d+1}{\alpha}+5dGD\right)
\left(2+\frac{\log\frac1{1-\lambda}}{\log K}\right)
+\frac2\alpha
\right]\log T,
\end{equation*}
simultaneously for all $\lambda\in(0,1)$.

When $K=2$, IFLH maintains $O(\log t)$ active experts and achieves
\begin{equation*}
\DReg(T,\lambda)=O\left(
d\left(\frac1\alpha+GD\right)
\log T\left(1+\log\frac1{1-\lambda}\right)
\right),
\end{equation*}
which matches the regret bound of AFLH. Alternatively, for any fixed integer $q\ge2$, choosing $K=\lceil T^{1/q}\rceil$ ensures $\lceil\log_K\tau\rceil\le q$ for all $\tau\in[T]$. Combining the strongly adaptive regret bound of IFLH with Corollary~\ref{cor:uniform}, we obtain
\begin{align*}
\DReg(T,\lambda)
&\le\left[
\frac{(5d+1)(q+1)+2}{\alpha}
+5d(q+1)GD
\right]\log T\\
&=O\left(
qd\left(\frac1\alpha+GD\right)\log T
\right),
\end{align*}
with $O(qT^{1/q})$ active experts.

%

\begin{remark}
For convex functions, CBCE matches the discounted regret bound of \citet{yang2026discounted} while extending the guarantee to all $\lambda\in(0,1)$. For strongly convex and exp-concave functions, IFLH with $K=2$ and AFLH achieve uniform discounted regret bounds with $O(\log t)$ active experts, whereas FLH improves the multiplicative logarithmic dependence to an additive dependence at the cost of $t$ active experts. Moreover, choosing $K=\lceil T^{1/q}\rceil$ in IFLH yields regret bounds independent of $\lambda$, with $O(qT^{1/q})$ active experts.
\end{remark}

\section{Conclusion}\label{sec:conclusion}

This paper demonstrates that strongly adaptive regret implies discounted regret uniformly over all discount factors. The reduction rests on an exact decomposition of the discounted regret against a fixed comparator into its regrets over suffixes of the horizon, and the mean suffix length under exponential discounting coincides with the effective window size. The decomposition applies to the decisions of any strongly adaptive method without modification, and existing methods thereby achieve uniform discounted regret bounds for convex, strongly convex, and exp-concave functions with their original computational cost. For convex functions, the resulting order matches that of the dedicated construction of \citet{yang2026discounted} on the interval considered there and extends to all discount factors in $(0,1)$.

\section*{Use of AI Tools}
During the development of the results, GPT-6 Pro was employed to explore proof approaches and refine the paper, including its writing and organization. The authors have verified all statements and proofs and take full responsibility for the content of this paper.

\vskip 0.2in
\bibliography{references}

\end{document}